\documentclass[times, review, 10pt]{elsarticle}
\usepackage{amssymb}
\usepackage{amsmath}
\usepackage[linesnumbered,ruled]{algorithm2e}
\usepackage{enumitem}
\usepackage{multirow}
\usepackage{diagbox}
\usepackage{booktabs}
\usepackage{wrapfig}
\usepackage[caption=false,font=normalsize,labelfont=sf,textfont=sf]{subfig}
\usepackage[utf8]{inputenc}
\usepackage{setspace}
\usepackage{pgfplots}
\pgfplotsset{compat=1.18}
\usepgfplotslibrary{fillbetween}
\usetikzlibrary{arrows.meta}
\definecolor{probA}{HTML}{2A78D6}
\definecolor{probB}{HTML}{EB6834}

\journal{}

\begin{document}

\begin{frontmatter}

\title{Distance to Class Prototypes: Active Learning for Object Detection}

\author[1]{Licheng Zhang\corref{cor1}}
\ead{licheng.zhang@student.unimelb.edu.au}
\cortext[cor1]{Corresponding author.}

\affiliation[1]{organization={School of Computing and Information Systems, The University of Melbourne},
            city={Melbourne},
            postcode={VIC 3010},
            country={Australia}}

\author[2]{Zheng Gong}
\ead{zheng.gong@jmu.edu.cn}

\affiliation[2]{organization={School of Computer Engineering, Jimei University},
            city={Xiamen},
            postcode={361021},
            state={Fujian},
            country={China}}

\begin{abstract}
Deploying a deep object detector in a new setting is limited less by architecture than by the cost of annotating data from that setting. Active learning lowers the cost by choosing which images to label, and the choice is only as good as the signal used to score an unlabeled image. That signal is usually the class posterior, which is cheap but poorly calibrated, or the disagreement across several models or several stochastic passes, which is better but multiplies inference over a pool far larger than the labeled set. We propose a signal richer than the posterior yet still read from one forward pass of one network. A supervised contrastive term added to the training objective shapes a per-object embedding space in which distance encodes class membership, and an unlabeled detection is scored by how far it lies from the region occupied by its predicted category, weighted by its confidence. The criterion needs no ensemble, no auxiliary predictor and no repeated inference, and its entire cost is 2.89M parameters, an increase of 8.3\% over a bare detector. On PASCAL VOC and MS-COCO it beats the posterior of the same detector in every round in which a selection is made, by up to 1.08\% mAP50 against run to run deviations of 0.02\% to 0.18\%, and it stays competitive with ensemble and Monte Carlo dropout criteria costing three to fifty forward passes per unlabeled image. Experiments use the single-stage detector under which the compared criteria report their results, so that the selection decision is isolated from the strength of the detector.
\end{abstract}

\begin{keyword}
Active learning \sep Object detection \sep Supervised contrastive learning \sep Informativeness function \sep Annotation cost \sep Representation learning
\end{keyword}

\end{frontmatter}

\section{Introduction}
\label{sec1}
The accuracy a deep object detector reaches in a new application is bounded in practice not by the availability of network architectures but by the availability of labeled data from the setting it must operate in. Each new deployment presents a distribution the detector has not seen, and a detector trained elsewhere must be adapted using images annotated for that setting. Annotation of a detection dataset is expensive, since a single image requires a set of tightly drawn bounding boxes instead of one categorical label, and the cost grows with every new domain the detector is asked to cover. Active learning addresses the cost directly by choosing which images are worth annotating \cite{joshi2009multi}.

The constraint that shapes the present work is that an annotation campaign run under a fixed compute budget cannot usually afford the machinery that the most accurate active learning methods require. An ensemble of detectors multiplies both the training cost and the memory that must be provisioned, and scoring the unlabeled pool by repeated stochastic inference is impractical when that pool is orders of magnitude larger than the labeled set and keeps growing. A selection criterion intended for such a setting must therefore be cheap by construction rather than cheap by approximation.

Concretely, active learning proceeds in cycles. A model is trained on a small labeled set, and is then used to assign every remaining unlabeled sample a score that estimates how much annotating it would improve the model. The highest scoring samples, and only those, are sent to a human annotator, the newly labeled samples are added to the labeled set, and the model is retrained. The cycle repeats until the annotation budget is exhausted or the target accuracy is reached. What distinguishes one active learning method from another is almost entirely the definition of that score, usually called the acquisition or informativeness function. The aim is to reach the accuracy of training on the fully labeled dataset while paying for a small fraction of the annotations. The setting is particularly attractive for object detection, where each annotation is correspondingly more expensive to obtain.

Initially, active learning was targeted at image classification \cite{ref1, ref16}. It was introduced to deep object detection in recent years \cite{ref9,ref12,ref3,ref10}. The methods that rank highest, however, tend to be the ones that pay most at selection time, and they do so in different ways that should not be conflated. Some enlarge the network or attach auxiliary predictors \cite{ref9, ref5}. Others leave the architecture entirely untouched but require the unlabeled pool to be scored several times over, whether by an ensemble of independently trained detectors \cite{ref6} or by repeated stochastic forward passes \cite{ref18}. Both are well founded ways to estimate uncertainty. Both also trade computation for accuracy, and the computation is spent on the unlabeled pool and not on the labeled set. Methods that avoid the cost, such as \cite{ref1, ref03}, generally score from the posterior alone and achieve weaker results. What is missing is a criterion that is richer than the posterior yet still obtainable from one pass of one network. That is the gap we address.

The cost framing determines which trade-offs count as favourable. A criterion that recovers the accuracy of an ensemble at a fraction of its cost is what an annotation campaign under a fixed compute budget requires, and every quantity we report, the parameter increment, the forward time and the per-image selection time, is stated in the units a practitioner would use to decide whether the method fits.

Contrastive learning builds a representation by embedding each anchor close to positive samples and away from negative ones, and has proved effective for downstream vision tasks \cite{simclr, moco}. In its self-supervised form a positive pair is an anchor and its augmented version, and a negative pair an anchor and a random sample from the minibatch \cite{simclr}. It has been brought to active learning by \cite{ref20}, which used contrastive coding to handle class distribution mismatch in the unlabeled pool. Supervised contrastive learning (SCL) \cite{ref021} uses the labels instead, so that a positive pair is two objects of the same category and a negative pair two objects of different categories.

\begin{figure}[t]
    \centering
    \includegraphics[width=0.8\linewidth]{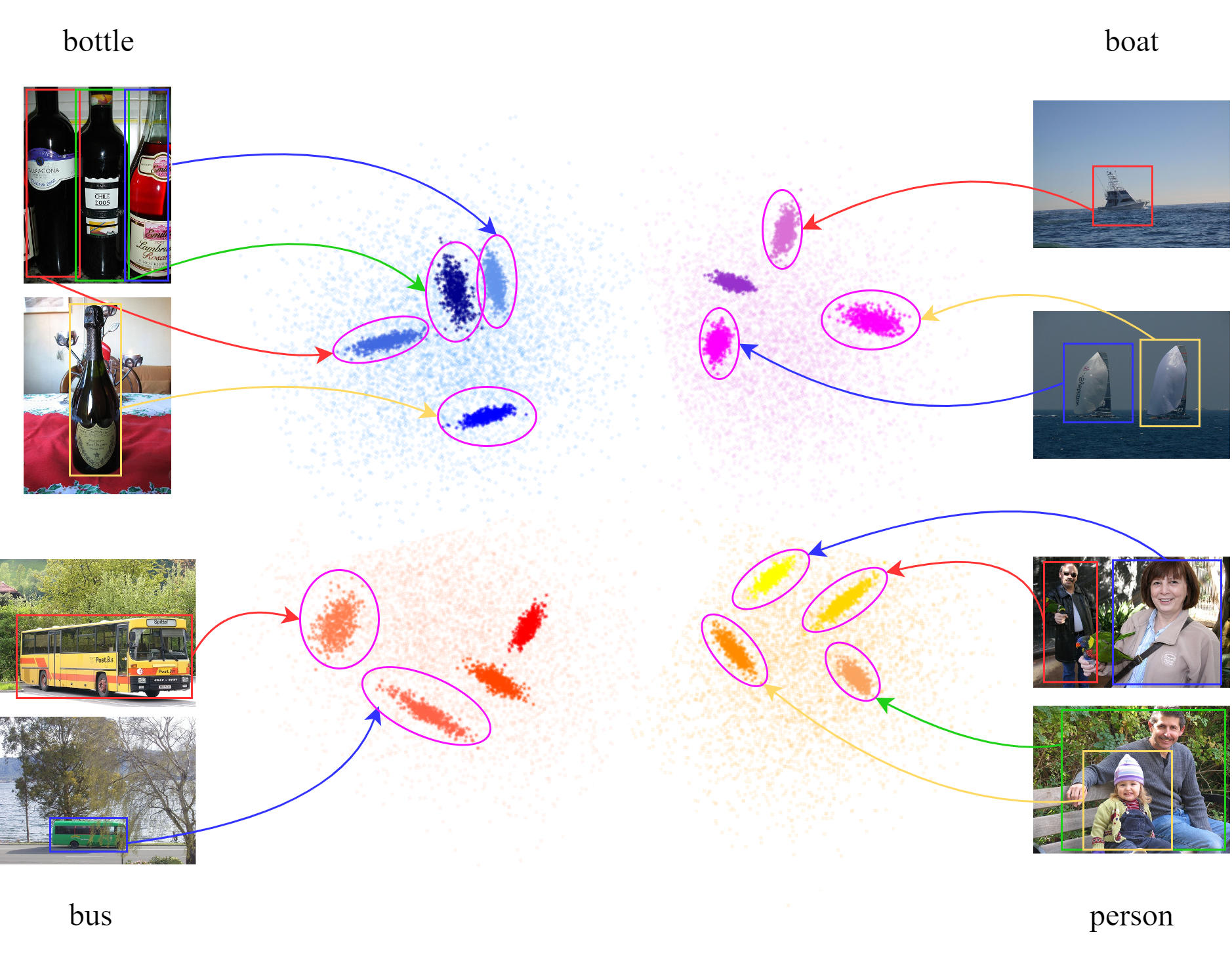}
    \caption{Contrastive embeddings of detected objects for four PASCAL VOC categories, one panel per category, the arrows linking sample detections to the group they fall in. Within a single category the embeddings split into several compact groups, ringed in magenta, each collecting one appearance of the object, the dark wine bottles against the champagne bottle or the motorboat against the two sailboats. A category therefore has no single centre, which is why our criterion scores a detection against the nearest anchor of its category rather than against a class mean.}
    \label{fig1}
\end{figure}

Inspired by the merit of SCL, we propose to leverage it for active learning of object detection. We incorporate it into the object detector and train the network with a summed loss of the detection objective and the supervised contrastive loss. To do so we attach an additional module, the contrastive module, to the detector, so that training pulls embeddings of the same class together and pushes those of different classes apart. Figure \ref{fig1} shows what the objective does to the embeddings of detected objects. Within a category they do not form a single blob but several compact sub-clusters, each collecting instances that look alike, and the scoring function we design reads that geometry to rank unlabeled images.

\subsection{Our contribution}
Our claim is not a faster or a more accurate detector, but a better signal to score unlabeled images with, established under a controlled comparison. The geometry of an embedding space shaped by the labels carries information about how informative a detection is that the class posterior of the same network does not, and it can be read from one forward pass. The comparison that establishes the point is with Entropy, which scores from the posterior of an identical detector under an identical protocol, and which our criterion beats in every round in which a selection is made, by up to 1.08\% mAP50 against run to run deviations of 0.02\% to 0.18\%. Against ensembles and Monte Carlo dropout, which cost three to fifty forward passes per image, our criterion is competitive while adding 2.89M parameters.

Ranking detections by the distance of their embeddings to those of labeled objects is not itself new, since Sokolov et al. \cite{sokolov25} select by cosine distance in a space obtained by principal component analysis. Our claim is about where the space comes from. To the best of our knowledge ours is the first work to shape it with a supervised contrastive objective for active learning of detection, so that distance is trained to encode class membership rather than to preserve variance. Section \ref{sec4.5} tests the distinction by holding the selection rule fixed and removing only the contrastive term. The modification to the detection head that makes the space available is described in Section \ref{sec3.2}. We regard it as an implementation detail rather than a contribution, being routine and specific to the detector we use.

Experiments cover three settings from PASCAL VOC and MS-COCO, with every criterion run on the same detector, pool and schedule, so that what is compared is the selection decision alone. They establish that the proposed signal is worth more than the posterior of the same network and that it remains competitive against criteria costing several times as much.

The rest of the paper is organized as follows. Section \ref{sec2} reviews the relevant literature, Section \ref{sec3} elaborates the proposed method, Section \ref{sec4} reports the experiments, Section \ref{sec5} states the limitations, and Section \ref{sec6} concludes.

\section{Related Work}
\label{sec2}
Systematic accounts of the area are given by Ren et al. \cite{ren21} for deep active learning at large, by Garcia et al. \cite{garcia23} for active learning of object detection specifically, and by Horchani \cite{horchani26}, who organises the field along acquisition logic, supervision granularity, operational regime and evaluation realism. We organise the discussion by the cost a method incurs, which is the axis we argue about.

\subsection{Criteria Computed Without Extra Networks}
An early line derives informativeness from quantities the detector already produces. Brust et al. \cite{ref2} built incremental learning with uncertainty-based detection metrics, Roy et al. \cite{ref03} used query-by-committee, and Kao et al. \cite{ref4} showed that localisation and classification uncertainty are both useful when sampling images. Wang et al. \cite{ref25} mined active samples adaptively for unseen categories, and Wang et al. \cite{ref30} compared least-confidence, margin and weighted classification sampling. A second group reduces annotation cost by changing what is queried rather than how it is scored. Desai et al. queried weak labels instead of boxes \cite{ref05} and sampled at box granularity \cite{ref06}, while Kothawade et al. \cite{ref020} mined targeted examples with submodular mutual information, and Wu et al. \cite{ref24} combined instance uncertainty with a diverse prototype strategy. Agarwal et al. \cite{ref8} measured contextual diversity.

\subsection{Criteria Requiring Additional Learning}
A second family attaches machinery to the detector. Yoo et al. \cite{ref5} predicted the loss of a sample with an auxiliary module, Aghdam et al. \cite{ref3} scored images from pixel-level detections, and Su et al. \cite{ref7} transferred representations across domains. Choi et al. \cite{ref9} estimated aleatoric and epistemic uncertainty with mixture density heads, Park et al. \cite{ref12} used evidential deep learning with hierarchical aggregation, and Yu et al. \cite{ref11} exploited consistency between box and class predictions. Yuan et al. \cite{ref10} modelled the relation between instance and image uncertainty, which Wan et al. \cite{midl23} later developed into a differentiation learning framework that unifies the instance and image levels, Liu et al. \cite{ref010} estimated unlabeled gradients, Li et al. \cite{ref09} studied aggregation over instances, and Tang et al. addressed transferability \cite{ref28} and calibration under architecture adaptation \cite{ref31}. Zhang et al. \cite{zhang} probed the robustness of detected objects by feature mixing, and Feng et al. \cite{ref34} proposed a common benchmark for the family. A related line trains a discriminator to separate labeled from unlabeled images, introduced by Sinha et al. \cite{vaal19} and made task aware by Kim et al. \cite{tavaal21}, which again pays for a second network trained alongside the detector.

\subsection{Recent Directions}
\label{sec2.2}
Four shifts since 2023 bear on the work reported here. The first is the pairing of uncertainty with an explicit diversity term, now close to standard. The pairing was established for classification by Ash et al. \cite{badge20}, who select by gradient embeddings that carry both quantities at once. In detection, Yang et al. \cite{ppal24} re-weight instance uncertainty by category difficulty and select with k-Means++, Hekimoglu et al. \cite{noris} filter redundant selections using features of detected regions, and Wang and Zhao \cite{umd25} add a class-aware prototype bank. Horchani \cite{horchani26} states the reason, that uncertainty is only a proxy for value. The second is portability across detectors. Here \cite{ppal24} and Sharma et al. \cite{pal26} change neither architecture nor training pipeline, the latter working purely from inference outputs, and Wen et al. \cite{asvp24} reuse pre-computed features from pre-trained models. The third re-examines what a query and a budget are. Zhang et al. \cite{delr} query at region rather than image level, Liang et al. \cite{mgral26} reward a reinforcement learning agent by mAP improvement, and Kassem Sbeyti et al. \cite{oss25} rank methods without training a detector, noting that one detector can cost up to 282 GPU hours. The fourth places acquisition inside domain adaptation rather than inside a single distribution. Menke et al. \cite{menke24} treat active selection as the bridge between unsupervised and supervised adaptation and propose strategies that account for the gap between a labeled source domain and an unlabeled target one. Their setting differs from ours in what the unlabeled pool represents, since their images are drawn from a distribution the detector was not trained on whereas ours share the distribution of the labeled set. Closest to us is Sokolov et al. \cite{sokolov25}, taken up in Section \ref{sec2.3}.

Active learning has also been applied to salient \cite{ref22, ref26}, aerial \cite{ref29, saood26} and UAV \cite{ref33} detection, and combined with semi-supervised \cite{ref22, ref27, ref32} and weakly supervised \cite{ref23} paradigms. Ensembles, which Haussmann et al. \cite{ref6} and Schmidt et al. \cite{ref21} scaled to production settings, and Monte Carlo dropout \cite{ref18} remain the strongest and most expensive baselines, and core-set selection \cite{ref1} the standard diversity-only reference.

\subsection{Positioning of Our Work}
\label{sec2.3}
The literature above is diverse in the signal it exploits, but it separates cleanly along the axis that matters in practice, namely what a method has to pay at selection time. Table \ref{tab:related} summarises the methods our experiments compare against along that axis, grouped as in Tables \ref{tab1} and \ref{tab2}, where Passes is the number of forward passes per unlabeled image required to score it, Extra network the additional network a method needs at selection time, and Parameters the count in millions under the protocol of Section \ref{sec4.4}. The more recent criteria of Section \ref{sec2.2} are reported on detectors for which these baselines give no results, so a selection-time cost stated under a different detector would not be commensurable with the column.

\begin{table}[t]
\caption{Selection-Time Cost of the Compared Methods on PASCAL VOC.}
\label{tab:related}
\footnotesize
\centering
\setlength{\tabcolsep}{2pt}
\begin{tabular}{l|l|c|l|r}
\toprule
Method & Signal & Passes & Extra network & Parameters\\
 & & & & (M)\\
\midrule
Entropy \cite{ref03} & posterior & 1 & none & 52.35\\
Core-set \cite{ref1} & feature coverage & 1 & none & 52.35\\
LLAL \cite{ref5} & predicted loss & 1 & loss module & 52.71\\
Prob \cite{ref9} & aleatoric and epistemic & 1 & probabilistic heads & 41.12\\
GMM \cite{ref9} & aleatoric and epistemic & 1 & GMM heads & 52.35\\
Feature-mixture \cite{zhang} & feature robustness & 1 & none & 35.02\\
\midrule
MC-dropout \cite{ref18} & posterior variance & 25--50 & none & 52.35\\
Ensemble \cite{ref6} & disagreement & 3 & two detectors & 157.05\\
\midrule
\textbf{Ours} & distance in SCL space & 1 & contrastive module & 37.91\\
\bottomrule
\end{tabular}
\end{table}

Two families are immediate from the table. The methods that rank highest buy accuracy with repeated inference over the entire unlabeled pool, the dominant cost in an annotation campaign. Those that keep a single forward pass either rely on the posterior alone or attach an auxiliary predictor whose training is a further source of variance. Ours stays in the single-pass regime and adds no auxiliary predictor, yet scores on a geometric property of a representation shaped by the labels themselves. Two recent lines deserve a direct comparison.

\textbf{Embedding distance criteria.} We are not the first to rank detections by embedding distance to labeled objects. Sokolov et al. \cite{sokolov25} do so using RoIAlign features reduced by principal component analysis, which preserves the directions of greatest variance and for a detector trained with cross entropy need not separate categories in any metric sense. A supervised contrastive objective optimises the metric itself, so distance is trained to mean class membership. Ours is the same functional form applied to a space in which the form is meaningful by construction, and Table \ref{tab:lossfunction} isolates exactly that factor by comparing the identical selection rule with and without $L_{sup}$.

\textbf{Detector-agnostic criteria.} A second line \cite{ppal24, pal26} changes nothing in the detector, scoring from inference outputs alone at zero architectural cost. Ours is the opposite trade, not a strictly better one. A criterion computed from outputs reads only what the posterior exposes, whereas a per-object embedding reveals how a detection sits relative to the population of objects the model has seen, at a cost of 2.89M parameters and one retraining. For a frozen third-party detector their choice is right, for a detector one trains oneself the cost is paid once and the signal is richer.

\textbf{Relation to our own earlier work.} Feature-mixture \cite{zhang}, which appears as a baseline in Tables \ref{tab1} and \ref{tab2}, is an earlier paper of the first author. We state the relationship instead of letting the shared authorship be discovered, since the two papers use the same detector, the same two benchmarks and the same protocol, and the reader is entitled to know what separates them. Feature-mixture scores a detected object by the robustness of its features under mixture with a set of base representations, which requires the features of every detection to be examined against every base representation, so its selection cost grows with the product of the two. The present criterion reads a geometric property of a space that a supervised contrastive term has trained, and its selection cost is one nearest anchor lookup against a fixed budget of $N_a$ anchors. The two therefore differ in what the score measures, in where the space being measured comes from, and in how the cost scales. They also differ in where they hold. On the twenty categories of PASCAL VOC the two are level at 3k and Feature-mixture is ahead by 0.20\% at 4k. On the eighty categories of MS-COCO, Feature-mixture reaches 28.60\% and 29.60\% against 28.67\% and 29.67\% for random selection, that is, it does not separate from random, whereas the present criterion reaches 29.00\% and 30.51\%. That is the comparison a reader should draw.

\section{Methodology}
\label{sec3}
\subsection{Overview}
We write ($X_L$, $Y_L$) for the labeled set, where $Y_L$ carries both the category and the bounding box of each object, and $X_U$ for the unlabeled pool. A detector is trained on ($X_L$, $Y_L$), the criterion of Section \ref{sec3.4} scores every image of $X_U$, and the $k$ highest scoring images are annotated and added to the labeled set before the detector is retrained.
\subsection{Contrastive Module}\label{sec3.2}
\begin{figure*}
\centering
\subfloat[]{\includegraphics[width=0.4\linewidth]{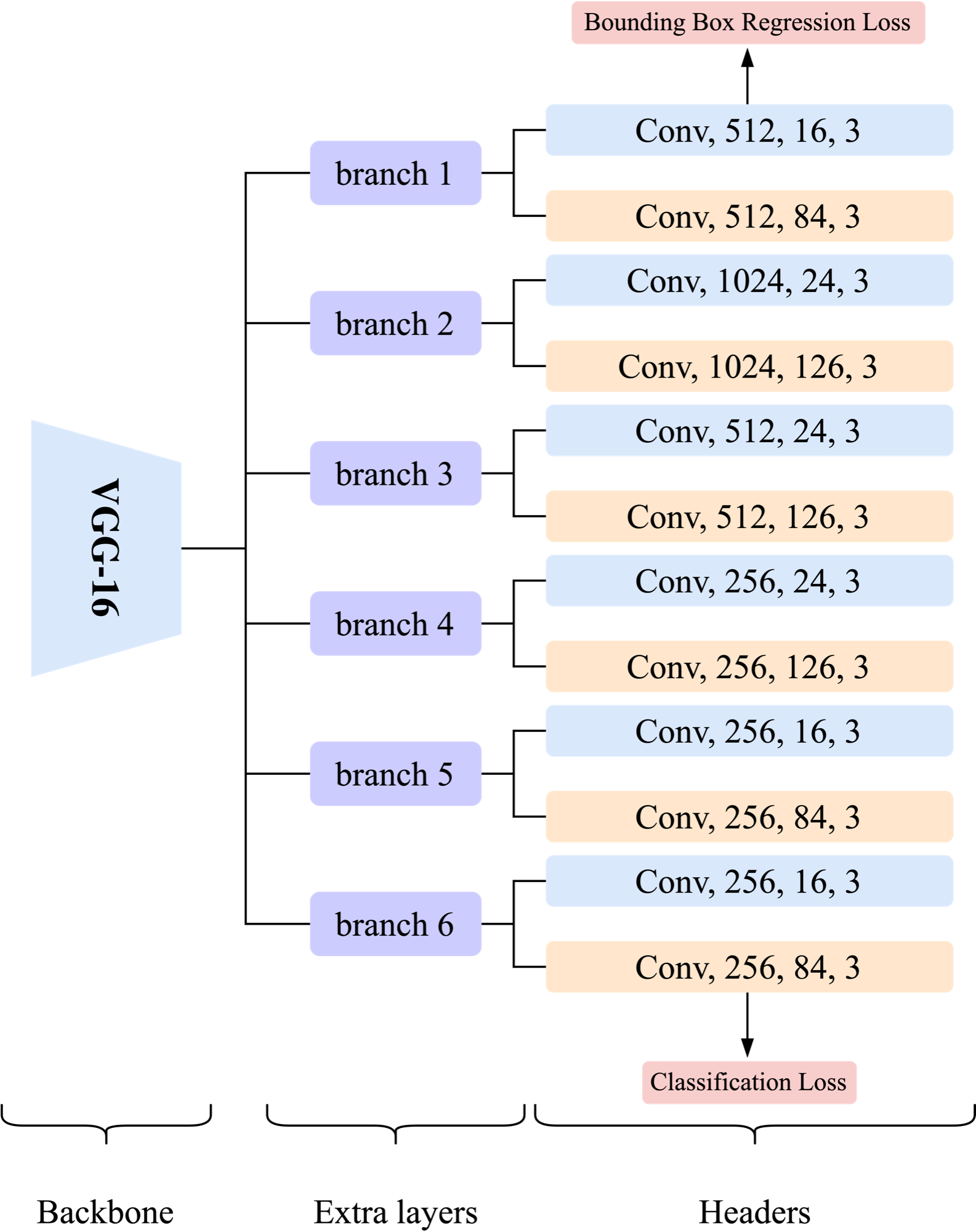}}
\hfill
\subfloat[]{\includegraphics[width=0.5\linewidth]{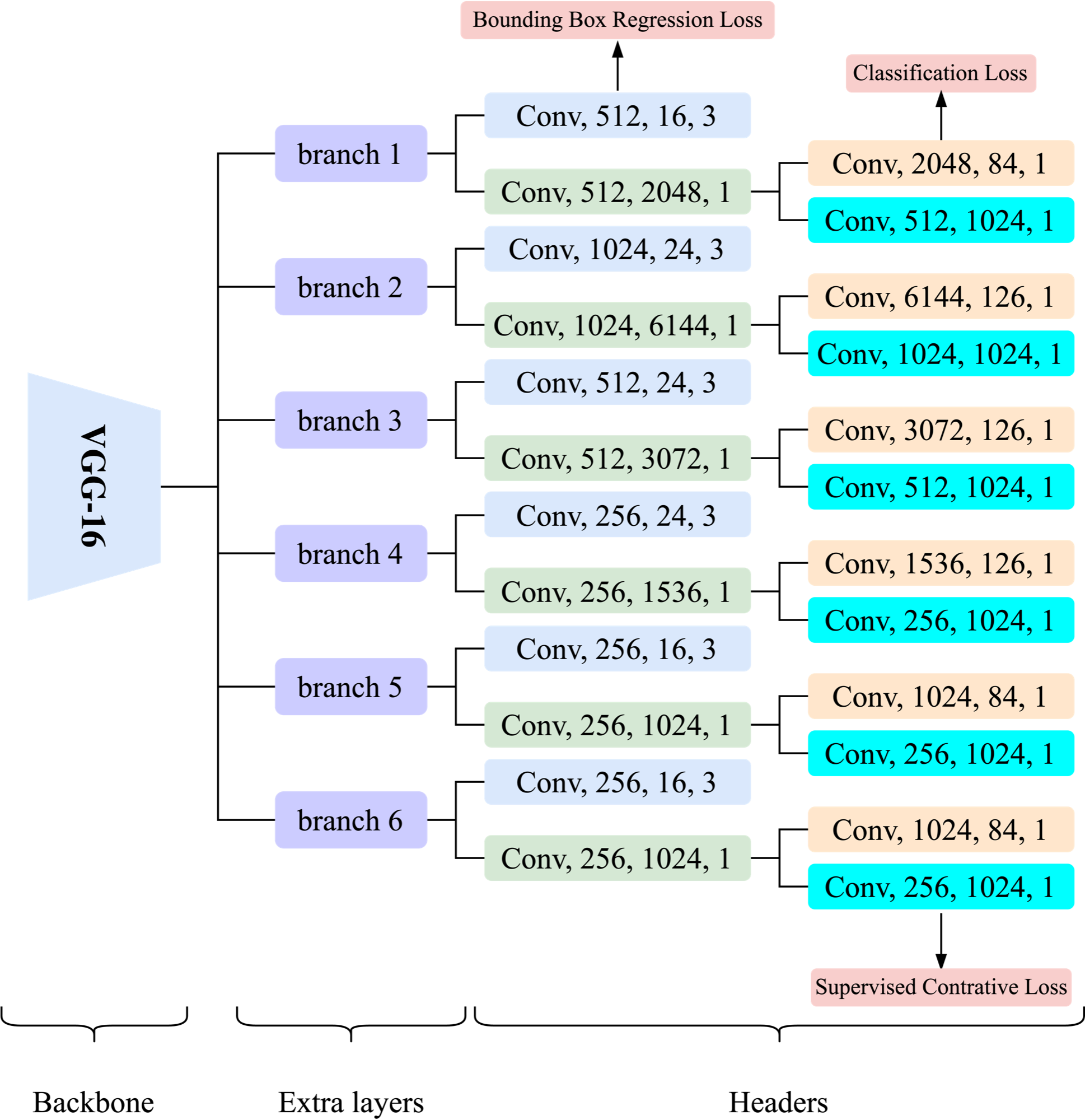}}
\caption{(a) The architecture of SSD \cite{ref13}. (b) Our adapted SSD.}
\label{fig:architecture}
\end{figure*}

We use SSD \cite{ref13}, a detector often utilized by state-of-the-art methods \cite{ref03,ref4,ref5,ref8,ref9,ref10,ref12} and therefore allows a direct comparison under an identical protocol. The only property our method requires of a detector is a layer at which a per-object embedding can be tapped and associated with a predicted category, which is available in dense detectors at the classification branch and in two-stage detectors at the region features, so the construction transfers in principle beyond SSD. SSD comprises a base network designed for image classification and auxiliary structures for detection. Figure \ref{fig:architecture}(a) illustrates its structure, in which no middle layer exposes the features of a detected object.

We make two adjustments to the architecture of SSD. Firstly, each classification layer is replaced by two consecutive convolutional layers, with the first one extracting features and the second one functioning as the classification layer. The input channels of the first layer as well as the output channels of the second one are the same as the original layer. The output channels of the first layer, \textit{i.e}, the input channels of the second layer, are set as follows.
\begin{equation}
  C_{o} = C_{i} \times n_b,
  \label{eq:o}
\end{equation}
In Equation \ref{eq:o}, $C_{i}$ is the number of input channels of the first layer and $n_b$ the number of default boxes per location. We use $n_b$ instead of the more common word `anchor' for the quantity, because in Section \ref{sec3.4} we reserve `anchor' for the reference embeddings $A_i^c$ that our criterion is defined against, and the two are unrelated. We employ the settings in \cite{ref13} by default, that is, $n_b$ is 4 and 6 for the two branches from the base network and 6, 6, 4 and 4 for the four branches from the auxiliary structures. To have a lower computational cost, we use a $1 \times 1$ kernel size for both layers. Figure \ref{fig:architecture}(b) illustrates the structure of the modified SSD. In Figure \ref{fig:architecture}, `Conv, $n_i$, $n_o$, $k$' denotes a convolutional layer with $n_i$ input channels, $n_o$ output channels and a $k \times k$ kernel. Light blue boxes are regression layers, light yellow boxes classification layers, and cyan boxes the layers added for SCL.

Secondly, to realize the training of SCL, we add another convolutional layer after the first convolutional layer in each branch, to embed objects' features to the same dimension. The number of output channels is set to 1024 and the kernel size is set to $1 \times 1$.
\subsection{SCL}
SCL is trained on the output of the last layer in Figure \ref{fig:architecture}(b), a 1024-dimensional normalized feature vector. The supervised contrastive loss \cite{ref021} is formulated as 
\begin{equation}
  L_{sup} = \sum_{i\in I}\frac{-1}{|P(i)|}\sum_{p\in P(i)}\log\frac{\exp(z_i\cdot z_p/\tau)}{\sum_{a\in A(i)}\exp(z_i\cdot z_a/\tau)},
  \label{eq:sup}
\end{equation}
In Equation \ref{eq:sup}, $P(i) \equiv \{p\in A(i) : y_p = y_i\}$ is the set of indices of all positives distinct from $i$. Unlike the original formulation of \cite{ref021}, where positives come from augmented views of the same image, here $I$ indexes the matched default boxes of a whole minibatch. A positive pair is two boxes of the same category, whether from the same image or from different ones, and no multi-view augmentation is involved. $A(i) \equiv I\backslash\{i\}$, where $I$ contains all samples in a minibatch. $|P(i)|$ is the cardinality of $P(i)$. $y_p$ and $y_i$ are labels. $z_i$, $z_p$ and $z_a$ are the contrastive features, which are normalized. $\tau$ is the scalar temperature. $\cdot$ denotes the inner product.

Following \cite{ref13}, we match default boxes to any ground truth box and take those with jaccard overlap above 0.5 as positive samples. Two features of the same category then form a positive pair and two of different categories a negative pair. Background, \textit{i.e.}, boxes whose jaccard overlap with every ground truth box is lower than 0.5, is ignored.

As a result, the whole loss function to train the modified SSD is formulated as
\begin{equation}
  L_{whole} = L_{conf} + L_{loc} + \lambda L_{sup},
  \label{eq:loss}
\end{equation}
where $L_{conf}$ and $L_{loc}$ are the confidence and localization loss respectively, which are the same as \cite{ref13}, and $\lambda$ balances the supervised contrastive term against the detection objective. We use $\lambda = 1$ throughout, that is, the three terms are summed with equal weight, and Section \ref{sec4.3} records the status of $\lambda$ and of the temperature $\tau$.
\subsection{Active Learning for Object Detection with SCL}
\label{sec3.4}
\textbf{Anchor construction.} SSD is first optimized on the labeled data with $L_{whole}$ until convergence. We then conduct inference on the labeled images to build the anchor set $A$, which serves as the class-wise reference, or prototype, of the feature space learnt by SCL. A category is represented by several anchors rather than by a single mean vector, for the reason set out under Equation \ref{eq:score}. Two design choices deserve explanation.

The first is why anchors are derived from detections rather than from ground truth boxes. The contrastive embedding of Section \ref{sec3.2} is produced per default box, so a ground truth box yields an embedding only after it has been matched to a default box by the jaccard criterion, as is done during training. Reusing the training-time matching at that stage would tie the anchor set to the matching hyperparameters instead of to what the detector predicts, and would produce embeddings for objects the detector cannot yet detect. We therefore run the ordinary inference path, and retain only objects whose confidence score exceeds 0.5, so that each anchor corresponds to a detection the model is already confident about. Because inference is carried out on the images the detector was trained on and only detections above 0.5 confidence are retained, the retained set is dominated by correct detections.

The second is the size of $A$. Retaining every qualifying detection would make the cosine search in Equation \ref{eq:score} grow with the size of the labeled pool, which would erode the efficiency that motivates our method. We therefore fix a total anchor budget $N_a$, drawing the anchors uniformly at random from the pool of qualifying detections, and discard categories for which no detection survives the threshold. All anchors are $\ell_2$ normalized. We set $N_a = 1000$ in all experiments, which is the budget under which the selection cost of Table \ref{tab:time} is measured. A detection on an unlabeled image can only carry a category the detector has learnt to predict. The anchor pool is built from the detections of that same detector on its training images, where the category is represented at least as well. Every predicted category we encountered was therefore present in the anchor set, and the maximum in Equation \ref{eq:score} was always taken over a non-empty set.

The budget matters less than it might appear, and the reason is visible in Equation \ref{eq:score}. The score of a detection depends on the anchor set only through $\mathop{max}\limits_{i} cos(P, A_i^c)$, that is, through the single nearest anchor of the predicted category. It is therefore a function of the outer surface of each class cluster, not of how densely that cluster is sampled. Once the anchors cover the directions the cluster occupies, drawing further anchors from its interior cannot change the maximum by more than the spacing between neighbouring anchors. That spacing shrinks quickly as the budget grows, because $L_{sup}$ concentrates the embeddings of one category by construction. A fixed budget in the low thousands is thus expected to behave like the uncapped set, while making the selection step independent of how large the labeled pool has become.

The construction is robust to the occasional incorrect detection that survives the threshold. Equation \ref{eq:score} takes a maximum over the anchor set, so such an anchor adds one further mode to a category without displacing the region it already occupies, and it can lower the score only of the detections that resemble it. The anchor set carries no learnable parameters either, so nothing in it propagates into the detector.

After that, we further perform inference on unlabeled images. Here the confidence threshold is lowered to 0.01, following the evaluation protocol of \cite{ref13}. The asymmetry with respect to the 0.5 threshold used above is deliberate. Anchors must be precise, since they define the reference directions of each class, whereas an unlabeled image must not be discarded merely because the detector is unconfident about it, which is exactly the regime active learning targets. The spurious boxes admitted by the low threshold are suppressed by the $prob$ factor in Equation \ref{eq:score} instead of by the threshold. Of the detections the detector returns we retain the top-$m$ by confidence to compute the informative score of the image. We set $m = 3$ in all experiments, so that both datasets are scored by an identical rule. A finite $m$ is necessary because the number of boxes surviving a 0.01 threshold varies by orders of magnitude across images, so averaging over all of them would let the score be dominated by how many boxes an image happens to produce. A small $m$ is necessary because the tail of the retained list consists of boxes whose $prob$ is near zero, which do not add information but enlarge the denominator of the mean and compress the scores of all images towards zero. A value of 3 avoids both failures and is of the order of the mean number of annotated objects per image in PASCAL VOC.

\begin{figure}[t]
    \centering
    \includegraphics[width=0.8\linewidth]{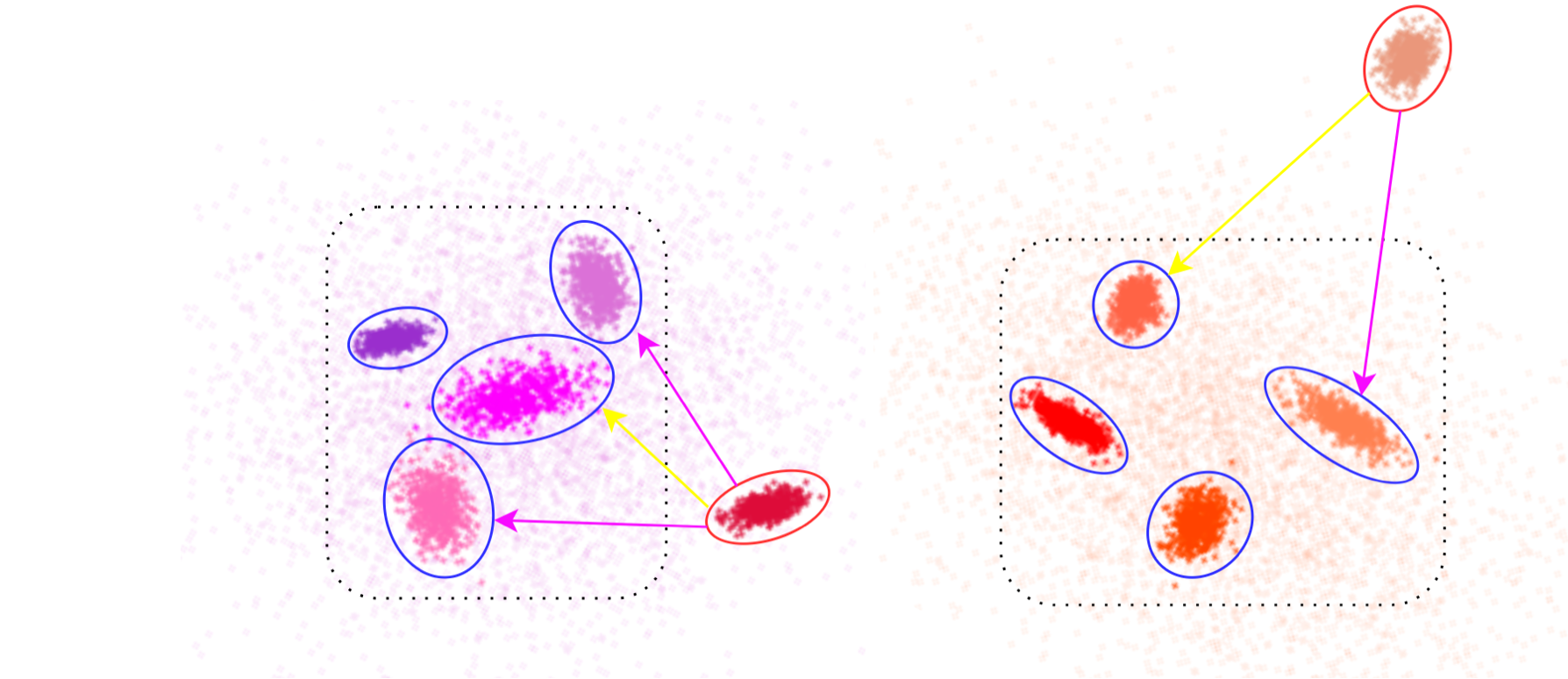}
    \caption{Scoring an unlabeled detection by its distance to the nearest anchor of its own predicted category.}
    \label{fig:distance}
\end{figure}

For each detection, we calculate its uncertainty using the following formula.
\begin{equation}
  uncer = (1 - \mathop{max}\limits_{i}cos(P, A_i^c)) \cdot prob,
  \label{eq:score}
\end{equation}
where $P$ is its normalized contrastive feature, $c$ is its category, and $prob$ is the confidence score. $A_i^c$ is $i$-th anchor of class $c$. $cos$ computes cosine similarity between $P$ and $A_i^c$. The uncertainty of a detection is carried entirely by the distance term $(1 - \mathop{max}\limits_{i}cos(P, A_i^c))$, which measures how far the detection lies from the region of the feature space occupied by its predicted category. The role of $prob$ is to weight that distance by how seriously the detection should be taken. The operation is a maximum over anchors rather than a distance to a class mean for the following reason. As Figure \ref{fig1} shows, a category does not occupy one compact region of the space but several, each collecting instances that resemble one another. The mean of a category can therefore fall in the empty space between its sub-clusters, reporting a large distance for a detection that is in fact typical of one of them. Taking the nearest anchor measures distance to the closest mode of the category instead of to its average, which is the quantity that matches what the embedding actually looks like. Under the 0.01 threshold the detector emits many spurious boxes, and such boxes are also far from every anchor, so without $prob$ they would dominate the image score and the selection would degenerate towards images full of detection noise. Weighting by confidence keeps the score driven by detections the model actually commits to, while still ranking a confidently predicted yet atypically embedded object as highly informative. Figure \ref{fig:distance} illustrates the mechanism of Equation \ref{eq:score}. In that figure the faint points are the contrastive embeddings of all detections of the predicted category $c$. The dotted box encloses the anchor set $\{A_i^c\}$ drawn from the labeled images, which the contrastive objective has concentrated into a small number of compact groups, each ringed in blue. The embedding $P$ of the detection being scored is ringed in red and lies outside the anchor set. The yellow arrow marks the nearest anchor group, whose similarity is the maximum in Equation \ref{eq:score} and the only one the score depends on. The magenta arrows mark other anchor groups of the same category, which are farther and do not enter the score. The detection on the right lies further from its category than the one on the left and is judged the more informative of the two, before confidence is taken into account.

The score of an image is the mean of the uncertainties of its retained detections, and the $k$ images of highest score, where $k$ is the collection size, are sent for annotation. Our active learning for object detection algorithm is summarized in Algorithm \ref{algorithm1}.
\begin{algorithm}[t]
\setstretch{1}\footnotesize
\caption{Active learning for object detection with SCL.}
\label{algorithm1}
\KwIn{Labeled data ($X_{L}$, $Y_{L}$), unlabeled data $X_{U}$, acquisition size $k$, total anchor budget $N_a$, detections per image $m$, modified SSD.}
\KwOut{$k$ most informative samples from $X_{U}$.}
Train SSD with ($X_{L}$, $Y_{L}$).\\
Perform inference on $X_{L}$ with threshold 0.5, and sample $N_a$ contrastive features uniformly at random to form the normalized anchor set $A$.\\
\textit{Data} = \{ \}.\\
\For{$x_u \in X_{U}$}{
Perform inference on $x_u$ with SSD.\\
Select top-$m$ detected objects with highest confidence scores and pick out their features $V$.\\
num = 0; value = 0.\\
\For{$v \in V$}{
Calculate $uncer$ for $v$ using Equation \ref{eq:score}.\\
num = num + 1; value = value + $uncer$.\\
}
\lIf{num $=$ 0}{score $=$ 0}
\lElse{score $=$ value\ /\ num}
Add score to \textit{Data}.
}
Select top-$k$ samples from \textit{Data}.
\end{algorithm}
\section{Experiments}
\label{sec4}
\subsection{Datasets}
We employed two popular object detection datasets for experiments: PASCAL VOC \cite{pascal} and MS-COCO \cite{coco}. PASCAL VOC consists of two subsets, VOC2007 (VOC07) and VOC2012 (VOC12). Following \cite{ref9, ref12, ref10}, for PASCAL VOC, we did experiments in two separate settings, 1) training and testing on VOC07, and 2) training on VOC07 and VOC12 (VOC07+12) and testing on VOC07. For MS-COCO, following \cite{ref9, ref12, ref10}, we trained on $train2014$ and evaluated on $val2017$ for a fair comparison.

Table \ref{tab:data} reports the statistics of the three settings, together with the active learning protocol applied to each. Training images are the \textit{trainval} split for VOC07, the union of the two \textit{trainval} splits for VOC07+12, and \textit{train2014} for MS-COCO. The two datasets differ in ways that matter for a criterion defined over per-object embeddings. MS-COCO has four times the number of categories and roughly three times as many annotated objects per image. An image therefore contributes more detections to the score, and each class prototype region is estimated from fewer anchors under a fixed budget. PASCAL VOC is the easier and more thoroughly reported setting, and MS-COCO the more demanding one.

\begin{table}[t]
\caption{Statistics of the Three Experimental Settings.}
\label{tab:data}
\small
\centering
\setlength{\tabcolsep}{4pt}
\begin{tabular}{l|c|c|c}
\toprule
 & VOC07 & VOC07+12 & MS-COCO\\
\midrule
Training images & 5011 & 16551 & 82783\\
Test images & 4952 & 4952 & 5000\\
Categories & 20 & 20 & 80\\
Initial labeled set & 2000 & 1000 & 5000\\
Samples per cycle & 1000 & 1000 & 1000\\
Number of cycles & 2 & 9 & 2\\
\bottomrule
\end{tabular}
\end{table}
\subsection{Experiment Settings}
Following \cite{ref9, ref12}, we employed the VGG-16 \cite{vgg} backbone for SSD, initialized using the pre-trained weight  on ImageNet \cite{imagenet}. For each experiment, SSD was trained for 500 epochs with 24 batch size and 50 warmup epochs, during which learning rate was linearly increased from $1e^{-5}$ to $1e^{-3}$, and after warmup, learning rate was set to $1e^{-3}$, which was decayed by 0.1 after 200 more and 100 more epochs respectively. mAP50 (IoU $>$ 0.5) was utilized as the evaluation metric, since it is the metric under which every baseline in Tables \ref{tab1} and \ref{tab2} reports its results, and adopting a stricter one would leave our numbers incomparable with the published results we place them beside. For PASCAL VOC, mAP50 was computed using VOC07 11 point method, and for MS-COCO, mAP50 was computed using MS-COCO evaluation method. The input resolution was uniformly $300 \times 300$. In the supervised contrastive loss, $\tau$ was set to 0.1. Optimisation used SGD with momentum 0.9 and weight decay $5 \times 10^{-4}$, and data augmentation followed \cite{ref13} unchanged, comprising photometric distortion, random expansion, random cropping by sampled jaccard overlap, and horizontal flipping. Both of our own arms, Ours and Random, were trained with exactly the recipe just given.

\subsection{Implementation Details}
\label{sec4.3}
The following details are what a reimplementation needs.

\textbf{Architecture.} Each of the six classification branches of SSD is replaced by two consecutive $1 \times 1$ convolutions. Denoting by $C_i$ the number of input channels of a branch and by $n_b$ its number of default boxes per location, the first convolution maps $C_i$ to $C_i \times n_b$ channels and the second maps $C_i \times n_b$ back to the original number of output channels. The values of $C_i$ are 512, 1024, 512, 256, 256 and 256 for the six branches, and the corresponding values of $n_b$ are 4, 6, 6, 6, 4 and 4, matching \cite{ref13}. The contrastive branch is attached in parallel to the input of the classification branch, a third $1 \times 1$ convolution mapping the $C_i$ input channels directly to 1024 channels, whose output is the contrastive embedding. The regression branches are untouched. The cost of the contrastive module is $\sum_j (C_i^{(j)} \cdot 1024 + 1024)$ over the six branches, which is $(512+1024+512+256+256+256) \cdot 1024 + 6 \cdot 1024 = 2{,}889{,}728$ parameters, the 2.89M reported throughout the paper.

\textbf{Training.} The contrastive embeddings of all six branches are reshaped to one list of 1024-dimensional vectors per image and $\ell_2$ normalised. A default box participates in $L_{sup}$ if it is matched to a ground truth box with jaccard overlap above 0.5, the same matching rule SSD already applies for $L_{conf}$ and $L_{loc}$, and its label is that of the matched ground truth box. Boxes matched to no ground truth box are excluded as background. $L_{sup}$ is accumulated over the whole minibatch rather than per image, so that positives and negatives are drawn across images. Equation \ref{eq:loss} is then optimised end to end from the ImageNet initialised backbone.

\textbf{Selection.} After training converges, inference is run on the labeled images with the confidence threshold at 0.5, and the contrastive embeddings of the surviving detections form the candidate anchor pool, from which $N_a = 1000$ are drawn uniformly at random. Inference is then run on the unlabeled images with the threshold at 0.01 and standard non maximum suppression. The $m = 3$ detections of highest confidence are retained per image, each is scored by Equation \ref{eq:score} against the anchors of its own predicted category, and the image score is their mean. An image from which the detector returns no box is assigned a score of zero. The $k$ images of highest score are queried. Only the contrastive embedding, the predicted category and the confidence of each retained detection are needed, so the selection step runs from a single cached inference pass over the unlabeled pool. None of $N_a$, $m$, $\lambda$ and $\tau$ was searched over.

\subsection{Experiment Results}
\label{sec4.4}
\textbf{Which baselines are compared, and why these.} The methods in Tables \ref{tab1} and \ref{tab2} are those that report results on SSD under the VOC and MS-COCO protocols we follow. An active learning comparison is only meaningful when every criterion selects from the same pool, for the same detector, under the same training schedule, because the quantity being compared is the value of a selection decision and not the strength of a detector. Recent criteria that are reported on stronger detectors cannot be transplanted into our protocol without either retraining them on SSD, which discards the tuning their authors performed and would misrepresent them, or upgrading our detector, which would break comparability with the entire body of results these benchmarks consist of. We chose comparability, and the consequence is a baseline set anchored to the SSD protocol. Section \ref{sec2} discusses the more recent methods \cite{ppal24, pal26, sokolov25, delr, mgral26} that we do not run for the reason just given, and Section \ref{sec5} records the resulting limitation. A reader should therefore take Tables \ref{tab1} and \ref{tab2} as evidence about selection criteria under a fixed detector, which is what they control for, and not as a claim about the current state-of-the-art in detection accuracy.

\textbf{Where the numbers come from.} Two rows of Tables \ref{tab1} and \ref{tab2} are run by us, namely Ours and Random, and every other row is quoted. Our own figures are the mean and standard deviation over three independent runs that differ only in the random seed, which controls the initial labeled subset and the network initialisation. The first round of each setting is the one exception and is reported as a mean alone, since no selection has been made at that point and the column compares no criterion against another. Random is run under the identical protocol and is the arm against which the cost of our contribution is measured below, which is why it is not quoted. On VOC07 the same runs also serve as the Random baseline of \cite{zhang}, so the two papers report the identical figure in that row. The quoted rows are taken from \cite{ref9} for every baseline except Feature-mixture, which is taken from \cite{zhang}, an earlier paper of the first author whose relationship to the present criterion is set out in Section \ref{sec2.3}.

We adopt the protocol of \cite{ref9} precisely so that the quotation is legitimate, and the correspondence is exact on every axis that governs an active learning result. The detector is SSD with a VGG-16 backbone initialised from ImageNet, the initial labeled set and the acquisition budget are those of Table \ref{tab:data}, the model is retrained from the ImageNet initialisation at every cycle rather than fine-tuned from the previous one, and the reported quantity is the mean and standard deviation of three independent trials.

Quoting compares against figures each method's own authors stand behind. What it does not give is a shared seed, so two rows start from the same initial labeled subset in distribution rather than identically, and differences of codebase enter the gap between rows alongside the difference in selection criterion. We therefore treat two methods as separated only when the gap exceeds the deviations involved, and the comparisons we rest on are those where it is several times that.

\textbf{Comparison with state-of-the-art on VOC07.} Random selection serves as the baseline, and we additionally train without the supervised contrastive loss.

Table \ref{tab1} reports the VOC07 results. We group the baselines by the selection-time cost of Table \ref{tab:related} and not by the family of the underlying algorithm, since cost is the axis we argue about. The first group scores an unlabeled image with one forward pass of one network, and comprises Entropy \cite{ref03}, Core-set \cite{ref1}, LLAL \cite{ref5}, Prob \cite{ref9}, GMM \cite{ref9} and Feature-mixture \cite{zhang}. We place Prob and GMM here, instead of among the heavier baselines as is sometimes done, because both are a single detector equipped with probabilistic prediction heads and neither requires repeated inference. The second group needs several forward passes or several networks, and comprises MC-dropout \cite{ref18} and Ensemble \cite{ref6}. In Tables \ref{tab1} and \ref{tab2}, bold marks the best result in a column, and results lying within one standard deviation of the best are bolded jointly. Table \ref{tab2} follows the same grouping and convention.

\textbf{How parameters are counted.} The parameter column reports what must be held in memory to run one selection step, which is what determines the deployment cost of an active learning system. It counts every network a method needs at selection time, not the trained detector alone. Each method is counted in the configuration under which its own authors report it, which is the configuration whose accuracy appears in the same row. The column is therefore the memory a practitioner would provision to reproduce each published method.

The claim we rest on is narrower and is measured like for like. Our method and the Random baseline are the same code path with and without the contrastive module of Section \ref{sec3.2}, trained and counted under one protocol, so the difference between them isolates the cost of our contribution exactly. That difference is 35.02M against 37.91M on PASCAL VOC and 39.88M against 42.77M on MS-COCO, which is 2.89M parameters on both datasets, an increase of 8.3\% and 7.2\% respectively over a bare detector. It is constant across the two datasets because the contrastive layer embeds to a fixed 1024 dimensions irrespective of the number of categories. The detector heads themselves grow with the category count, 21 for VOC against 81 for MS-COCO, which is why the two bare detectors differ. Table \ref{tab:time} shows the corresponding statement in time, a forward pass of 0.0037s against 0.0032s. Where a genuinely like for like comparison against a competitor is available it points the same way. Ensemble \cite{ref6} is three independently trained detectors, hence three times the memory, three times the training budget and three forward passes per unlabeled image, and MC-dropout \cite{ref18} pays 25 to 50 forward passes per unlabeled image at the same memory as one detector. Neither of those multipliers depends on how the underlying detector is configured, and both dwarf the 8.3\% we add.

\begin{table}[t]
\caption{Comparative Results on VOC07 With State-of-the-Art. Ours and Random are run by us, the remaining rows quoted from \cite{ref9} and, for Feature-mixture, from \cite{zhang}. $^{\dagger}$The first round carries no selection and is given as a mean alone, for the reason in Section \ref{sec4.4}.}
\label{tab1}
\scriptsize
\centering
\setlength{\tabcolsep}{1pt}
\begin{tabular}{l|c|c|c|c}
\toprule
\multirow{2}{*}{Method} & \multicolumn{3}{c|}{mAP50 (\%)} & Parameters\\
& 1st (2k)$^{\dagger}$ & 2nd (3k) & 3rd (4k) & (M)\\
\midrule
Random & 62.37 & 66.34$\pm$0.17 ($\uparrow$ 3.97) & 68.39$\pm$0.69 ($\uparrow$ 2.05) & 35.02\\
Entropy \cite{ref03} & 62.43 & 66.85$\pm$0.12 ($\uparrow$ 4.42) & 68.70$\pm$0.18 ($\uparrow$ 1.85) & 52.35\\
Core-set \cite{ref1} & 62.43 & 66.57$\pm$0.20 ($\uparrow$ 4.14) & 68.57$\pm$0.26 ($\uparrow$ 2.00) & 52.35\\
LLAL \cite{ref5} & 62.47 & 67.02$\pm$0.11 ($\uparrow$ 4.55) & 68.90$\pm$0.15 ($\uparrow$ 1.88) & 52.71\\
Prob \cite{ref9} & 62.91 & 67.61$\pm$0.17 ($\uparrow$ 4.70) & 69.66$\pm$0.17 ($\uparrow$ 2.05) & 41.12\\
GMM \cite{ref9} & 62.43 & 67.32$\pm$0.12 ($\uparrow$ 4.89) & 69.43$\pm$0.11 ($\uparrow$ 2.11) & 52.35\\
Feature-mixture \cite{zhang} & 62.37 & \textbf{67.76$\pm$0.03} ($\uparrow$ \textbf{5.39}) & \textbf{69.98$\pm$0.01} ($\uparrow$ 2.22) & 35.02\\
\midrule
MC-dropout \cite{ref18} & 62.43 & 67.10$\pm$0.07 ($\uparrow$ 4.67) & 69.39$\pm$0.09 ($\uparrow$ \textbf{2.29}) & 52.35\\
Ensemble \cite{ref6} & 62.43 & 67.11$\pm$0.26 ($\uparrow$ 4.68) & 69.26$\pm$0.14 ($\uparrow$ 2.15) & 157.05\\
\midrule
\textbf{Ours} & 62.50 & \textbf{67.77$\pm$0.02} ($\uparrow$ 5.27) & 69.78$\pm$0.02 ($\uparrow$ 2.01) & 37.91\\
\bottomrule
\end{tabular}
\end{table}

Within the single forward pass group our method and Feature-mixture \cite{zhang} are on par at 3k, differing by 0.01\% against standard deviations of 0.02\% and 0.03\%, and at 4k Feature-mixture is ahead by 0.20\%. Our method is ahead of every other member of the group at both rounds. Against the second group, which pays several forward passes per unlabeled image, our method is better at both rounds.

\textbf{The main comparison against the posterior of the same detector.} Before turning to the remaining benchmarks we isolate the comparison the paper is about. Entropy \cite{ref03} scores an unlabeled image from the class posterior. The figures we quote come from \cite{ref9}, which runs it on the same detector and backbone, from an initial set of the same size, with the same acquisition budget and the same retraining schedule. The comparison controls the protocol, which is what governs an active learning result, but not the codebase, so the gap carries the value of the signal together with whatever implementation difference separates the two. The ablation of Section \ref{sec4.5} controls both, since both of its arms are our own code, and it points the same way. On VOC07 the gap is 0.92\% at 3k and 1.08\% at 4k, and on MS-COCO, reported below, it is 0.07\% at 6k and 0.62\% at 7k. Three of the four gaps are an order of magnitude above the run to run deviations, which are 0.02\% to 0.18\% in these rows, and all four are positive. The first round of each setting is not part of the comparison, since no selection has been made at that point and both criteria train on the same random initial subset. The difference there, $+0.07$\% on VOC07 and $-0.37$\% on MS-COCO, reflects the effect of $L_{sup}$ on detector training rather than the value of a selection signal. Our claim is accordingly stated over the rounds in which a selection is made. The distance term therefore carries information about which images are worth annotating that the posterior of the same detector under the same protocol does not expose, which is our claim.

Our score is a product of the distance term and $prob$, so the comparison establishes that the posterior \emph{augmented by} the distance term beats the posterior alone, not that the distance term alone would. Figure \ref{fig:prob} shows the distance term used without $prob$, which is weaker than the full score, so the geometry adds to the posterior rather than replacing it.

The comparison with $S_{dis}$ in Table \ref{tab:lossfunction} makes the complementary point, that the information is not recovered by an arbitrary function of the same embeddings. The column that removes $L_{sup}$ shows that it is not present in the embeddings the detector produces without the contrastive objective. Taken together the three comparisons locate the contribution precisely. It is neither the distance rule alone nor the embeddings alone, but the pairing of a rule that reads geometry with a space in which geometry has been trained to mean something.

\begin{table}[t]
\caption{Comparative Results on MS-COCO With State-of-the-Art. Conventions as in Table \ref{tab1}.}
\label{tab2}
\scriptsize
\centering
\setlength{\tabcolsep}{1pt}
\begin{tabular}{l|c|c|c|c}
\toprule
\multirow{2}{*}{Method} & \multicolumn{3}{c|}{mAP50 (\%)} & Parameters\\
& 1st (5k)$^{\dagger}$ & 2nd (6k) & 3rd (7k) & (M)\\
\midrule
Random & 27.17  & 28.67$\pm$0.12 ($\uparrow$ 1.50) & 29.67$\pm$0.09 ($\uparrow$ 1.00) & 39.88\\
Entropy \cite{ref03} & 27.70 & 28.93$\pm$0.11 ($\uparrow$ 1.23) & 29.89$\pm$0.09 ($\uparrow$ 0.96) & 116.51\\
Core-set \cite{ref1} & 27.70 & 28.99$\pm$0.01 ($\uparrow$ 1.29) & 29.93$\pm$0.06 ($\uparrow$ 0.94) & 116.51\\
LLAL \cite{ref5} & 27.71 & 28.71$\pm$0.06 ($\uparrow$ 1.00) & 29.53$\pm$0.15 ($\uparrow$ 0.82) & 116.87\\
Prob \cite{ref9} & 27.33 & 29.06$\pm$0.08 ($\uparrow$ \textbf{1.73}) & 30.02$\pm$0.05 ($\uparrow$ 0.96) & 73.20\\
GMM \cite{ref9} & 27.70 & \textbf{29.28$\pm$0.05} ($\uparrow$ 1.58) & \textbf{30.51$\pm$0.12} ($\uparrow$ 1.23) & 116.51\\
Feature-mixture \cite{zhang} & 27.10 & 28.60$\pm$0.00 ($\uparrow$ 1.50) & 29.60$\pm$0.00 ($\uparrow$ 1.00) & 39.88\\
\midrule
MC-dropout \cite{ref18} & 27.70 & 29.20$\pm$0.09 ($\uparrow$ 1.50) & 30.30$\pm$0.08 ($\uparrow$ 1.10) & 116.51\\
Ensemble \cite{ref6} & 27.70 & 29.03$\pm$0.07 ($\uparrow$ 1.33) & 30.02$\pm$0.06 ($\uparrow$ 0.99) & 349.53\\
\midrule
\textbf{Ours} & 27.33 & 29.00$\pm$0.01 ($\uparrow$ 1.67) & \textbf{30.51$\pm$0.01} ($\uparrow$ \textbf{1.51}) & 42.77\\
\bottomrule
\end{tabular}
\end{table}

\textbf{Comparison with state-of-the-art on MS-COCO.} Table \ref{tab2} reports the MS-COCO results. At 7k our method reaches 30.51\%, which is the best result in the table, matched only by GMM \cite{ref9} at the same value but with 2.7 times the parameters. Our gain over the preceding round, 1.51\%, is the largest of any method, and the gap is well outside the standard deviations involved. At 6k our improvement over the initial round ranks second, 1.67\% against the best 1.73\%, while the absolute figure of 29.00\% is behind the 29.28\% of GMM and the 29.20\% of MC-dropout. The comparison with Feature-mixture \cite{zhang} is the one that separates the two datasets. On the twenty categories of VOC07 the two criteria are level, whereas on the eighty categories here Feature-mixture reaches 28.60\% and 29.60\% against the 28.67\% and 29.67\% of random selection, so it returns no gain over choosing images at random, while ours reaches 29.00\% and 30.51\%. Section \ref{sec2.3} sets out what differs between the two criteria. The parameter comparison holds here too, our 42.77M being second only to Feature-mixture and rather less than half the 116.51M shared by most of the table.

\begin{figure*}[t]
\centering
\subfloat[]{\includegraphics[width=0.48\linewidth]{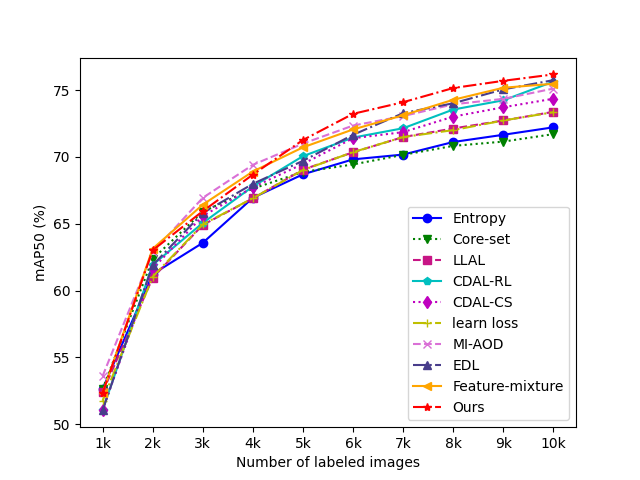}}
\hfill
\subfloat[]{\includegraphics[width=0.48\linewidth]{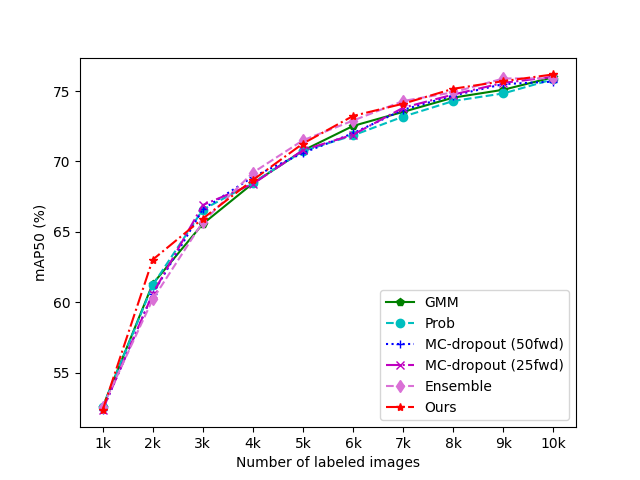}}
\caption{\textbf{VOC07+12.} Comparison against methods that score with (a) a single pass of a single detector and (b) the remaining baselines. Baseline curves are quoted from their sources rather than re-run, on the same basis as Tables \ref{tab1} and \ref{tab2}.}
\label{fig:voc0712}
\end{figure*}

\textbf{Comparison with state-of-the-art on VOC07+12.} We compared our method with state-of-the-art single-model based methods, including Entropy \cite{ref03}, Core-set \cite{ref1}, LLAL \cite{ref5}, CDAL-RL \cite{ref8}, CDAL-CS \cite{ref8}, learn loss \cite{ref5}, MI-AOD \cite{ref10}, EDL \cite{ref12} and Feature-mixture \cite{zhang}. Figure \ref{fig:voc0712}(a) illustrates the results. Nine rounds by ten methods does not fit a column-width table, so the setting is reported as curves and the comparisons drawn from it are ordinal. Our method is above every one of them except MI-AOD and Feature-mixture at all nine rounds. Against MI-AOD the curves cross, with MI-AOD ahead at 3k and 4k, and against Feature-mixture they cross at 2k, 3k and 4k, ours being ahead at the remaining rounds in both cases. The numerical comparisons we rest on are made in Tables \ref{tab1} and \ref{tab2}, where deviations accompany every mean.

We further compared our method with the remaining baselines, namely GMM \cite{ref9}, Prob \cite{ref9}, MC-dropout at 50 and at 25 passes \cite{ref18}, labelled 50fwd and 25fwd in the figure, and Ensemble \cite{ref6}. Prob and GMM \cite{ref9} appear in panel (b) for continuity with how they are plotted in the prior literature, although by the criterion of Tables \ref{tab1} and \ref{tab2} they belong with the single pass methods. Our method is ahead of GMM in all nine rounds, ahead of Prob and of MC-dropout at 25 passes in all but 3k, and ahead of MC-dropout at 50 passes in all but 3k and 4k. Against Ensemble it is ahead at 2k, 3k, 6k, 8k and 10k and behind at 4k, 5k, 7k and 9k. Ensemble, however, holds four times the parameters at selection time, which is what decides affordability at the scale an annotation campaign runs at. Read together with Table \ref{tab:related}, our method stays level with criteria costing several times as much while holding a fraction of their parameters.

\begin{table}[t]
\caption{Comparison of Computational Cost.}
\label{tab:time}
\footnotesize
\centering
\setlength{\tabcolsep}{2pt}
\begin{tabular}{c|c|c|c}
\toprule
\multirow{2}{*}{Step} & \multirow{2}{*}{SSD \cite{ref13}} & \multirow{2}{*}{Our SSD} & Scoring \\
& forward & forward & (1000 anchors)\\
\midrule
Time (second) & 0.0032 & 0.0037 & 0.008\\
\bottomrule
\end{tabular}
\end{table}

\textbf{Forward time analysis.} We measured the running time of our network, of the original SSD \cite{ref13}, and of the selection step itself, on an NVIDIA GeForce RTX 4090 GPU. Table \ref{tab:time} presents their results. The active learning column reports the per-image cost of scoring with \eqref{eq:score} at the budget $N_a = 1000$ of Section \ref{sec3.4}. The adapted SSD increases the forward time from 0.0032s to 0.0037s, and scoring one unlabeled image costs a further 0.008s.

Scoring an unlabeled image with our criterion costs one forward pass of the adapted detector plus one nearest anchor lookup, 0.0037s plus 0.008s, hence 0.0117s in total. MC-dropout costs 25 or 50 stochastic passes of the unmodified detector, hence 0.08s or 0.16s at the measured 0.0032s per pass, which is 6.8 or 13.7 times ours. Ensemble costs three passes, 0.0096s, so the two are comparable per image and what separates them is memory, 157.05M parameters resident at selection time against our 37.91M, which is what decides on an embedded platform.

The two components of our cost scale differently across the pool. The forward pass is strictly linear in the number of unlabeled images, since no part of it is shared between them. The scoring step multiplies the retained embeddings of an image by the same $1000 \times 1024$ anchor matrix for every image, so it is a single matrix product with one fixed operand and batches across the pool, and the 0.008s above is measured one image at a time and carries the overhead of an individual call. What multiplies in MC-dropout and in Ensemble is the component that cannot be shared.

The figures for both competing methods are the measured cost of one pass of the unmodified detector multiplied by the pass count their definitions fix. They exclude the repeated model loading and the aggregation those methods perform, so they are a lower bound on the wall clock cost.
\subsection{Ablation Studies}
\label{sec4.5}
We ablate here the two components our method consists of, namely the supervised contrastive term that shapes the embedding space and the scoring function that reads it, each by substitution rather than by removal of a hyperparameter. Both are reported in the single Table \ref{tab:lossfunction}, whose columns differ in exactly one component while everything else, including the detector, the schedule and the seeds, is held fixed. In that table, `Without $L_{sup}$' trains the identical network without the supervised contrastive term and selects with the identical rule, which isolates the shaped embedding space. `$S_{dis}$' keeps our training and replaces our scoring function with the one proposed in \cite{ref20} for image classification, which isolates the scoring function. A dash marks a configuration we did not run. That rule assumes one embedding per image, so transplanting it to detection requires a choice of how to aggregate over the detections of an image, and we ran it on the two settings where that choice is least contestable. Section \ref{sec4.3} records the quantities that are hyperparameters and not components, none of which was searched over.

\textbf{Supervised contrastive loss.} The first column removes $L_{sup}$ from Equation \ref{eq:loss} and leaves everything else in place, so the network trains on the detection objective alone and selection still uses Equation \ref{eq:score}, now read off the space the detector produces on its own. The comparison is run on all three settings and on every round of each. The full method is ahead in each of the nine rounds, which is what the shaped space contributes over the unshaped one under an identical selection rule. The paragraph that closes the subsection gives the largest of the gaps.

\begin{table}[t]
\caption{Ablation of the Two Components of Our Approach.}
\label{tab:lossfunction}
\footnotesize
\centering
\setlength{\tabcolsep}{2pt}
\begin{tabular}{c|c|c|c|c}
\toprule
\multirow{2}{*}{\textbf{Dataset}} & \multirow{2}{*}{\textbf{Samples}} & \multicolumn{3}{c}{mAP50 (\%)} \\
& & Without $L_{sup}$ & $S_{dis}$ \cite{ref20} & \textbf{Ours} \\
\midrule
\multirow{3}{*}{\textbf{VOC07}} & 1st (2k) & 62.37 & -- & \textbf{62.50} \\
& 2nd (3k) & 67.59 & 67.48 & \textbf{67.77} \\
& 3rd (4k) & 69.25 & -- & \textbf{69.78} \\
\midrule
\multirow{3}{*}{\textbf{MS-COCO}} & 1st (5k) & 27.17 & -- & \textbf{27.33} \\
& 2nd (6k) & 28.77 & -- & \textbf{29.00} \\
& 3rd (7k) & 29.57 & -- & \textbf{30.51} \\
\midrule
\multirow{3}{*}{\textbf{VOC07+12}} & 2nd (2k) & 62.70 & 62.99 & \textbf{63.02} \\
& 3rd (3k) & 65.83 & -- & \textbf{65.94} \\
& 4th (4k) & 68.20 & -- & \textbf{68.68} \\
\bottomrule
\end{tabular}
\end{table}

\textbf{Multiplied by $prob$.} The confidence term $prob$ in Equation \ref{eq:score} plays a vital role in our scoring function. We therefore repeated the experiments with $prob$ removed from the scoring function. Experiments were performed on VOC07+12 from 5k to 10k, that is, over the last six of the ten rounds of that schedule.

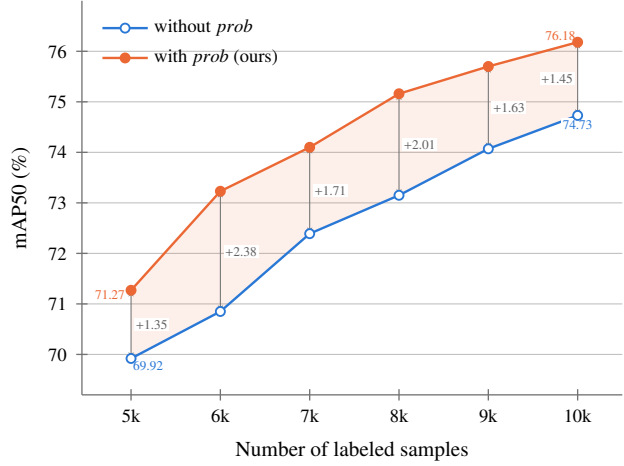
\begin{figure}[t]
    \centering
\begin{tikzpicture}
\begin{axis}[
    width=0.72\textwidth, height=0.56\textwidth,
    axis lines=left,
    xmin=-0.55, xmax=5.5, ymin=69.2, ymax=77.0,
    enlarge x limits=false, enlarge y limits=false,
    xtick={0,1,2,3,4,5},
    xticklabels={5k,6k,7k,8k,9k,10k},
    ytick={70,71,72,73,74,75,76},
    xlabel={Number of labeled samples},
    ylabel={mAP50 (\%)},
    every axis x label/.style={at={(ticklabel cs:0.5)}, anchor=near ticklabel, font=\footnotesize},
    every axis y label/.style={at={(ticklabel cs:0.5)}, rotate=90, anchor=near ticklabel, font=\footnotesize},
    tick label style={font=\scriptsize},
    ymajorgrids=true,
    grid style={black!25, line width=0.4pt},
    axis line style={-, black!55, line width=0.5pt},
    tick style={black!55, line width=0.5pt},
    legend style={at={(0.02,0.98)}, anchor=north west, draw=none,
                  fill=none, font=\scriptsize, row sep=1pt},
    legend cell align=left,
    clip=false,
    /tikz/gaparrow/.style={{Stealth[length=2.4pt,width=2pt]}-%
                           {Stealth[length=2.4pt,width=2pt]},
                           black!55, line width=0.4pt},
]
\addplot[name path=lo, draw=none, forget plot]
    coordinates {(0,69.92) (1,70.85) (2,72.39) (3,73.15) (4,74.07) (5,74.73)};
\addplot[name path=hi, draw=none, forget plot]
    coordinates {(0,71.27) (1,73.23) (2,74.10) (3,75.16) (4,75.70) (5,76.18)};
\addplot[probB, opacity=0.10, forget plot] fill between[of=lo and hi];

\addplot[probA, line width=0.9pt, mark=*, mark size=1.7pt,
         mark options={fill=white, draw=probA, line width=0.7pt}]
    coordinates {(0,69.92) (1,70.85) (2,72.39) (3,73.15) (4,74.07) (5,74.73)};
\addlegendentry{without $prob$}
\addplot[probB, line width=0.9pt, mark=*, mark size=1.7pt,
         mark options={fill=probB, draw=probB, line width=0.7pt}]
    coordinates {(0,71.27) (1,73.23) (2,74.10) (3,75.16) (4,75.70) (5,76.18)};
\addlegendentry{with $prob$ (ours)}

\draw[gaparrow] (axis cs:0,69.92) -- (axis cs:0,71.27);
\node[anchor=west, font=\tiny, text=black!62, fill=white, inner sep=0.6pt,
      xshift=1pt] at (axis cs:0,70.595) {+1.35};
\draw[gaparrow] (axis cs:1,70.85) -- (axis cs:1,73.23);
\node[anchor=west, font=\tiny, text=black!62, fill=white, inner sep=0.6pt,
      xshift=1pt] at (axis cs:1,72.040) {+2.38};
\draw[gaparrow] (axis cs:2,72.39) -- (axis cs:2,74.10);
\node[anchor=west, font=\tiny, text=black!62, fill=white, inner sep=0.6pt,
      xshift=1pt] at (axis cs:2,73.245) {+1.71};
\draw[gaparrow] (axis cs:3,73.15) -- (axis cs:3,75.16);
\node[anchor=west, font=\tiny, text=black!62, fill=white, inner sep=0.6pt,
      xshift=1pt] at (axis cs:3,74.155) {+2.01};
\draw[gaparrow] (axis cs:4,74.07) -- (axis cs:4,75.70);
\node[anchor=west, font=\tiny, text=black!62, fill=white, inner sep=0.6pt,
      xshift=1pt] at (axis cs:4,74.885) {+1.63};
\draw[gaparrow] (axis cs:5,74.73) -- (axis cs:5,76.18);
\node[anchor=east, font=\tiny, text=black!62, fill=white, inner sep=0.6pt,
      xshift=-1pt] at (axis cs:5,75.455) {+1.45};

\node[anchor=north west, font=\tiny, text=probA, fill=white, inner sep=0.6pt]
    at (axis cs:0,69.92) {69.92};
\node[anchor=north, font=\tiny, text=probA, fill=white, inner sep=0.6pt,
      yshift=-1pt] at (axis cs:5,74.73) {74.73};
\node[anchor=north east, font=\tiny, text=probB, fill=white, inner sep=0.6pt,
      xshift=-1.5pt, yshift=1pt] at (axis cs:0,71.27) {71.27};
\node[anchor=south east, font=\tiny, text=probB, fill=white, inner sep=0.6pt]
    at (axis cs:5,76.18) {76.18};
\end{axis}
\end{tikzpicture}
    \caption{\textbf{VOC07+12.} Effect of the $prob$ factor in Equation \ref{eq:score}.}
    \label{fig:prob}
\end{figure}

In Figure \ref{fig:prob}, `with $prob$' denotes the scoring function multiplied by $prob$ and `without $prob$' the variant lacking it. The full score is ahead in every one of the six rounds, by 1.35\%, 2.38\%, 1.71\%, 2.01\%, 1.63\% and 1.45\% from 5k to 10k. Each gap is an order of magnitude above the run to run deviations of these experiments, and the margin does not close as the labeled set grows, so the confidence weighting is not a correction that only matters while labels are scarce. The mechanism is the one set out under Equation \ref{eq:score} in Section \ref{sec3.4}, and the figure is the measurement of it. Six rounds covering the second half of the schedule settle the question the ablation asks. The factor multiplies the score of every retained detection, so it would have to reverse its effect partway through one schedule for the earlier rounds to disagree, and the measured margin neither decays nor trends towards closing across the six rounds shown.

\textbf{Scoring function.} The third column of Table \ref{tab:lossfunction} isolates the second component. Here the network is trained exactly as ours, so the embedding space is the same, but selection uses the scoring function $1 - (cos_{st} - cos_{nd} + cos_{st, nd})$ proposed in \cite{ref20} for active learning of image classification in place of Equation \ref{eq:score}. Our scoring function is ahead in both settings where the comparison was run, by 0.29\% on VOC07 and by 0.03\% on VOC07+12.

Read together, the two columns say something more informative than either alone. On VOC07 at 3k, removing $L_{sup}$ costs 0.18\% while swapping the scoring function costs 0.29\%, and on VOC07+12 at 2k the figures are 0.32\% and 0.03\%. Neither component dominates, so the result of Section \ref{sec4.4} rests on the pairing rather than on either alone. Across the table every one of the nine cells is in favour of the full method and none reverses, and the largest gaps, 0.94\% on MS-COCO at 7k, 0.53\% on VOC07 at 4k and 0.48\% on VOC07+12 at 4k, all fall in the `Without $L_{sup}$' column. The shaped space therefore carries the effect, since the identical selection rule performs worse when applied to the space the detector produces without $L_{sup}$. That is the comparison relevant to the relationship with \cite{sokolov25} discussed in Section \ref{sec2.3}.

\subsection{Visualization and analysis}
\begin{figure*}[t]
    \centering
    \includegraphics[width=0.95\linewidth]{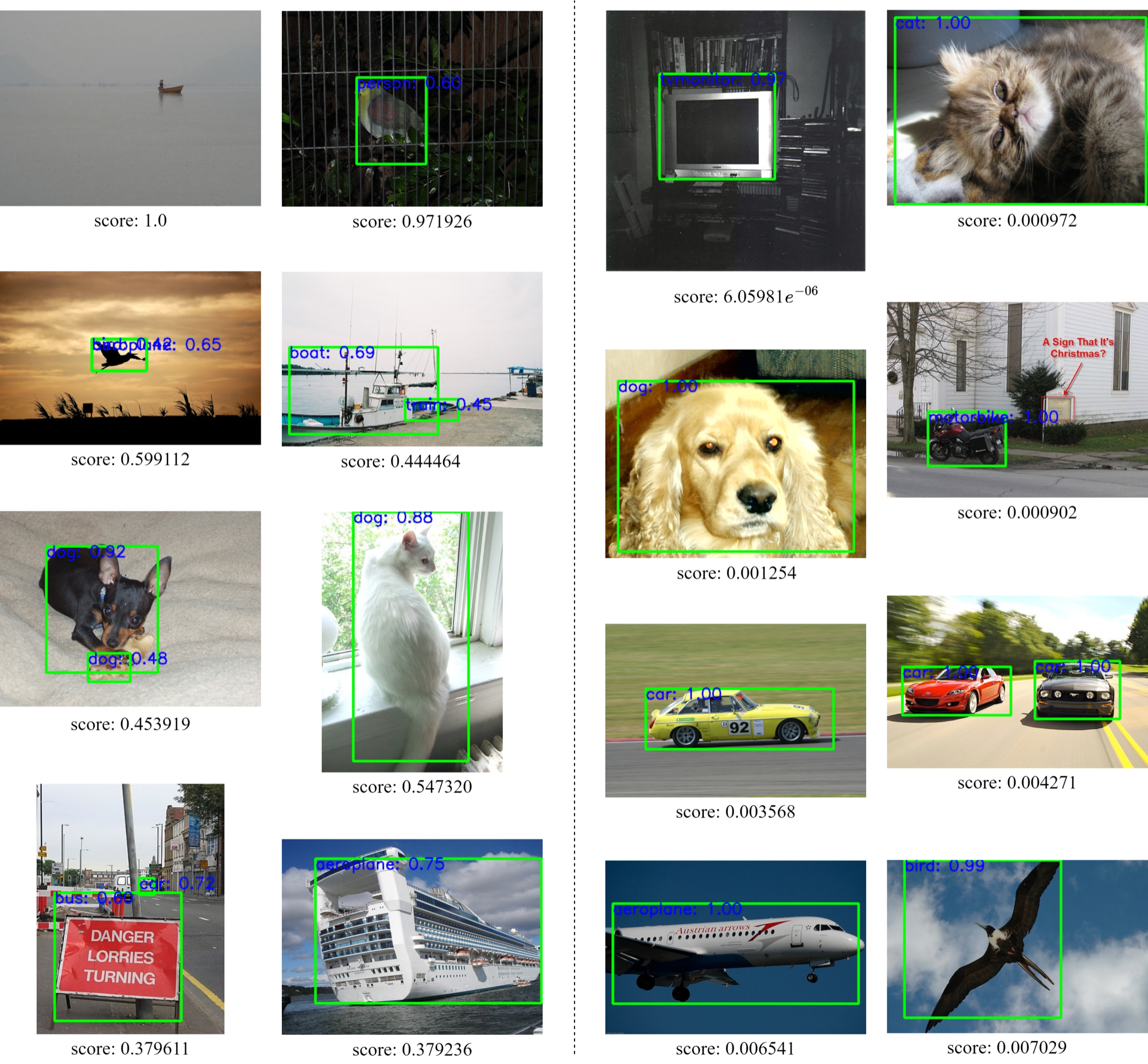}
    \caption{Selected samples. The eight on the left have high informative scores, those on the right low.}
    \label{fig3}
\end{figure*}

Figure \ref{fig3} shows examples selected in the first round on VOC07+12, drawn with a display threshold of 0.4 and at most ten boxes per image. The eight on the left received high informative scores and the eight on the right low ones. Every image on the left carries a detection the model has got wrong, a bird predicted as a person, one object assigned two classes, a part of an object detected as a dog, a cat detected as a dog, a boat detected as an aeroplane, or no object above the display threshold at all. Every image on the right is detected accurately. The criterion therefore assigns high scores to the images the detector is getting wrong.

\textbf{Failure modes.} The same figure exposes where the criterion should be expected to fail, and the structure of Equation \ref{eq:score} predicts three distinct cases. The first follows directly from the limitation discussed in Section \ref{sec5}. A detection whose category is correct, whose confidence is high and whose embedding is typical will receive a low score no matter how poorly its box is placed. An image containing only badly localised instances of easy categories is therefore invisible to our criterion, while a criterion that models localisation uncertainty such as \cite{ref4} would select it. The second concerns objects of a category absent from the labeled set. Such an object is necessarily assigned to some known category, and if it is assigned confidently and its embedding happens to fall near that category's anchors, it receives a low score, which is the opposite of what is wanted. The criterion measures atypicality with respect to the categories the model already knows, and cannot by construction measure novelty with respect to categories it does not. The third is a consequence of averaging over $m$ detections. An image containing one highly informative object among several routine ones has its score diluted towards that of the routine ones, so our criterion prefers images that are uniformly difficult over images that are difficult in one place.
\section{Limitations and Future Work}
\label{sec5}
Four limitations bound the result.

\textbf{The criterion is defined on the classification branch.} No term in Equation \ref{eq:score} depends on where a box is or on how well it is placed, so a detection that is confidently and typically embedded while badly localised will be scored as uninformative. The criterion addresses the recognition component of detection, and stratifying contrastive pairs by jaccard overlap rather than by category alone is the most direct route to extending it to localisation, which methods such as \cite{ref4, ref6} treat explicitly.

\textbf{There is no diversity term.} The criterion is purely one of uncertainty, so a single acquisition round can select redundant images, and a line of work from \cite{badge20} onwards pairs the two for that reason \cite{ppal24, noris, umd25, horchani26}. Our embedding makes the component cheap to add. A redundancy filter or a clustering step would operate on the contrastive features Equation \ref{eq:score} has already computed, requiring no further forward pass. The pairing is therefore a natural extension of the work and not a competing design.

\textbf{One detector at one resolution.} All our experiments use a single-stage detector at $300 \times 300$, which is the configuration under which the compared baselines report their results, but which leaves the criterion unverified on the heavier detectors used when the inference budget is not binding. The backbone initialisation was likewise held fixed at ImageNet-pretrained VGG-16. Sensitivity to that choice should be limited, since $L_{sup}$ optimises the geometry our score reads instead of inheriting it, unlike a method scoring in a space it does not train such as \cite{sokolov25}.

\textbf{Semi-supervised detection is not compared.} Semi-supervised detection targets the same regime by extracting signal from the unlabeled images rather than by choosing which of them to annotate. The two are complementary and have been combined \cite{ref22, ref27, ref32, ref23}, so nothing here should be read as a claim that selecting samples is preferable to exploiting unlabeled data.

Beyond detection, the construction assumes only per-instance predictions carrying a category and an embedding, which instance segmentation and multi-label classification satisfy directly and pure regression does not.

\section{Conclusion}
\label{sec6}
We have argued that the geometry of a label-shaped embedding space is a better signal for active learning of object detection than the class posterior of the same network, and that it can be read at the cost of one forward pass. We shape the space with a supervised contrastive objective, score an unlabeled detection by its distance from the region occupied by its predicted category, and establish the claim against Entropy, which reads the posterior of an identical detector under an identical protocol and which our criterion beats in every round in which a selection is made, by up to 1.08\% mAP50. Two further comparisons locate the effect, showing that neither the distance rule alone nor the embeddings alone account for it. Against criteria that pay three to fifty forward passes per unlabeled image, our criterion is competitive while adding 2.89M parameters and one nearest anchor lookup, a margin that matters when the unlabeled pool is orders of magnitude larger than the labeled set. Section \ref{sec5} sets out the limitations that bound the result.
\section*{CRediT authorship contribution statement}
\textbf{Licheng Zhang:} Conceptualization, Methodology, Software, Investigation, Visualization, Writing -- original draft. \textbf{Zheng Gong:} Supervision, Writing -- review and editing.

\section*{Declaration of competing interest}
The authors declare that they have no known competing financial interests or personal relationships that could have appeared to influence the work reported here.

\section*{Acknowledgements}
Partial financial support for this study was provided by the National Natural Science Foundation of China (NSFC) through grants 42301468 and 42371457, the Fujian Province Natural Science Foundation, China (Grant No. 2023J01799), and the Xiamen Natural Science Foundation, China (Grant No. 3502Z20227048), along with contributions from the Jimei University Startup Fund (Grant ZQ2022031).

\bibliographystyle{elsarticle-num-names}
\bibliography{els.bib}

~\\~\\~\\
\begin{wrapfigure}{l}{25mm}
\includegraphics[width=1in,height=1.25in,clip,keepaspectratio]{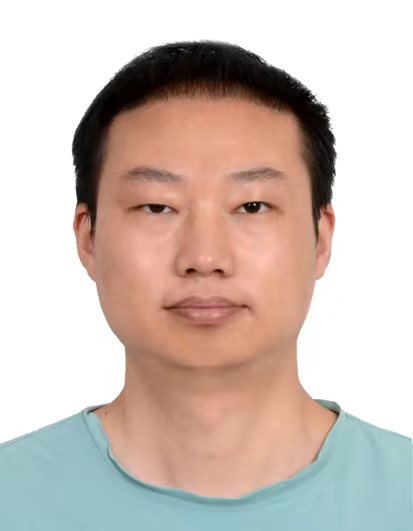}
\end{wrapfigure}\par
\textbf{Licheng Zhang} graduated from Yangzhou University in 2013 with a bachelor degree in Information and Computing Science. He graduated from Peking University in 2016 with a master degree in Intelligence Science and Technology. He is currently pursuing the Ph.D. degree with The University of Melbourne, Melbourne, VIC, Australia. His research interests lie in computer vision, deep learning and pattern recognition.

~\\~\\~\\
\begin{wrapfigure}{l}{25mm}
\includegraphics[width=1in,height=1.25in,clip,keepaspectratio]{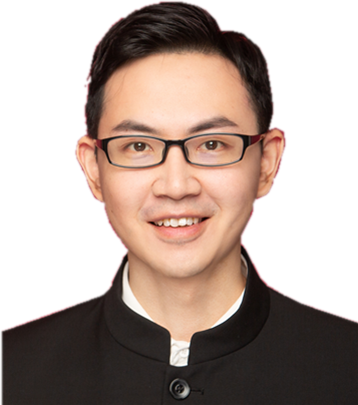}
\end{wrapfigure}\par
\textbf{Zheng Gong} received his Ph.D. in Communication and Information Systems from Xiamen University and was a visiting Ph.D. scholar at the University of Waterloo, Canada. He is currently a Research Fellow and Lecturer at the College of Computer Engineering, Jimei University, Xiamen. Previously, he worked as a Visual Algorithm Researcher at DJI Innovations in Shenzhen. His main research interests include computer vision and 3D perception for robotics, covering areas such as multi-sensor state estimation and optimization, point cloud semantic analysis, online calibration of multi-sensor fusion, and integrated perception-control networks based on reinforcement learning.

\end{document}